\documentclass[letterpaper]{article} 
\usepackage{aaai25}  
\usepackage{times}  
\usepackage{helvet}  
\usepackage{courier}  
\usepackage[hyphens]{url}  
\usepackage{graphicx} 
\usepackage{natbib}  
\usepackage{caption} 
\usepackage{algorithm}
\usepackage{algorithmic}

\usepackage{newfloat}
\usepackage{listings}
\DeclareCaptionStyle{ruled}{labelfont=normalfont,labelsep=colon,strut=off} 
\floatstyle{ruled}
\newfloat{listing}{tb}{lst}{}
\floatname{listing}{Listing}
\title{Ambiguity-Restrained Text-Video Representation Learning \\for Partially Relevant Video Retrieval}
\author{
    Cheol-Ho Cho, WonJun Moon, Woojin Jun, MinSeok Jung, and Jae-Pil Heo\thanks{Corresponding author}
}
\affiliations{
    Sungkyunkwan University\\

    \{hoonchcho, wjun0830, junwoojinjin, minseokjung0328, jaepilheo\}@gmail.com
}

\usepackage{bibentry}

\usepackage{amsmath}
\usepackage{amssymb}
\usepackage{arydshln}
\usepackage{array}
\usepackage{hhline}
\usepackage{boldline} 
\usepackage{colortbl}
\usepackage{xcolor}
\usepackage{multirow}
\usepackage{multicol}
\usepackage{enumitem} 
\usepackage{lipsum} 
\usepackage{booktabs}

\begin{document}

\maketitle

\begin{abstract}
Partially Relevant Video Retrieval~(PRVR) aims to retrieve a video where a specific segment is relevant to a given text query.
Typical training processes of PRVR assume a one-to-one relationship where each text query is relevant to only one video.
However, we point out the inherent ambiguity between text and video content based on their conceptual scope and propose a framework that incorporates this ambiguity into the model learning process.
Specifically, we propose Ambiguity-Restrained representation Learning~(ARL) to address ambiguous text-video pairs.
Initially, ARL detects ambiguous pairs based on two criteria: uncertainty and similarity.
Uncertainty represents whether instances include commonly shared context across the dataset, while similarity indicates pair-wise semantic overlap.
Then, with the detected ambiguous pairs, our ARL hierarchically learns the semantic relationship via multi-positive contrastive learning and dual triplet margin loss.
Additionally, we delve into fine-grained relationships within the video instances.
Unlike typical training at the text-video level, where pairwise information is provided, we address the inherent ambiguity within frames of the same untrimmed video, which often contains multiple contexts. 
This allows us to further enhance learning at the text-frame level.
Lastly, we propose cross-model ambiguity detection to mitigate the error propagation that occurs when a single model is employed to detect ambiguous pairs for its training.
With all components combined, our proposed method demonstrates its effectiveness in PRVR.

\end{abstract}

\begin{figure}[t]
    \begin{center}
        \noindent\includegraphics[height=2.1in, width=0.48\textwidth]{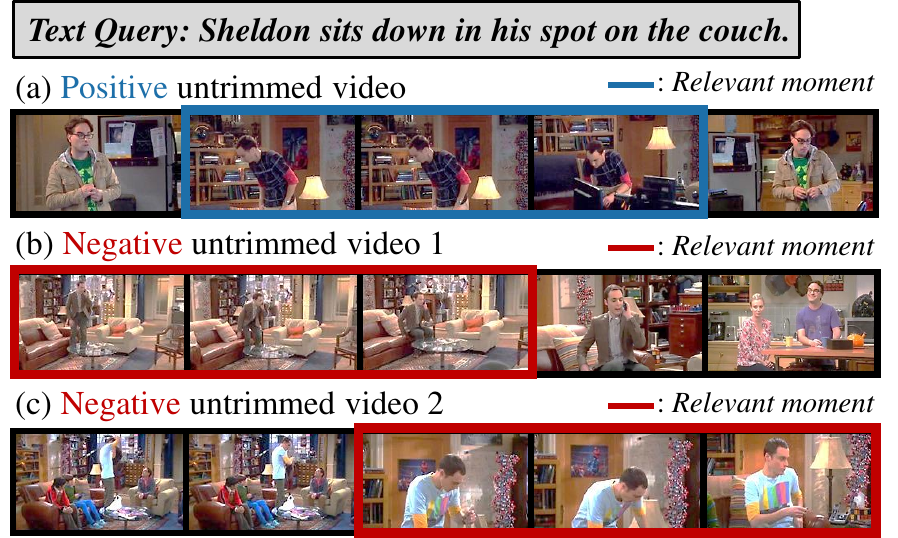}
    \end{center}    
    \caption{
    The illustration of ambiguous relationships between text-video pair. For the text query, \textit{`Sheldon sits down in his spot on the couch.'}, all three untrimmed videos contain relevant scenes. However, only video (a) is learned as positive, while videos (b) and (c) are typically treated as negative in prior techniques. Such ambiguous relationships between text and video are more likely to occur with untrimmed videos that often include diverse contexts.
    }
    \label{fig:motivation}
\end{figure}

\section{Introduction}
\label{sec:intro}
With societal advancements, the use of video media for information dissemination has become widespread. 
Consequently, the field of Text-to-Video Retrieval (T2VR), which allows users to find desired videos using text queries, has also been spotlighted. 
However, existing T2VR approaches often assume that videos contain only the relevant parts to the text query. 
This assumption does not align with real-world scenarios, where videos can vary in length and context. 
To address this issue, partially relevant video retrieval~(PRVR)~\cite{ms-sl} is proposed to deal with untrimmed videos where only specific video segments correspond to the text query. 

MS-SL~\cite{ms-sl} initially modeled multi-scale video features to prepare for various contexts that a text query might encompass. 
GMMFormer~\cite{wang2024gmmformer}, on the other hand, critiqued the exhaustive multi-scale clip modeling of MS-SL for its inefficiency and proposed to use Gaussian attention to encode only the local contexts.
Although these methods have made significant progress in PRVR, the obscurity in labeling text-video pairs is a yet-to-be-explored problem.

Typically, text and video instances are labeled pair-wise due to the high costs of exploring relationships between all text and video instances.
This has led previous works to learn only the paired instance as positive relations, treating all others as negative, even when similar video-text pairs are present.
However, we argue that pair-wisely labeled retrieval datasets often introduce ambiguity between text-video instances. 
For example, as illustrated in Fig.~\ref{fig:motivation}, while the text query is clearly partially relevant to the paired video at the top, it is also relevant to other video instances in the dataset.

\begin{figure}[t]
    \begin{center}
        \noindent\includegraphics[height=1.85in, width=0.46\textwidth]{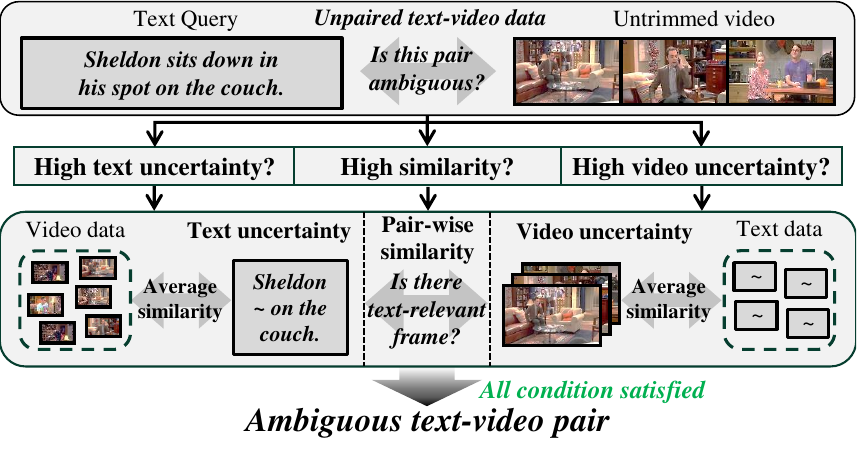}
    \end{center}    
    \caption{
        Illustration for ambiguous pair detection.
        To identify ambiguous text-video pairs, we use two key metrics: uncertainty and similarity. 
        Uncertainty for each text and video instance is calculated by measuring the average similarity across different modalities.
        This reflects the degree of contextual overlap within the dataset.
        Similarity is determined by the maximum similarity between the text and frames from the untrimmed video.
        We define a pair as having an ambiguous relationship when both uncertainty and similarity are high, exhibiting the commonly shared context in the dataset and between the pair.
    }
    \label{fig:high_level_method}
\end{figure}

In this regard, we propose Ambiguity-Restrained representation Learning (ARL), a framework that detects instances in ambiguous relations with the model's online~(per-epoch) knowledge.
These ambiguities are taken into account in the objectives to reduce fallible supervision which can arise when all unpaired pairs are treated as having a negative relationship.
To determine ambiguous relationships, we use uncertainty and similarity measures, as shown in Fig.\ref{fig:high_level_method}.
Simply put, we define the text-video pairs in ambiguous relations if they exhibit high uncertainty and high pair-wise similarity, indicating that the pair cannot be simply defined as negative since they possess commonly shared semantics throughout the dataset and their resemblance to each other.
Our proposed ARL includes these identified ambiguous relationships in training.
Particularly, we grant the model the flexibility in treating instances with ambiguous relationships by relaxing the constraints while positive and negative relationships are learned in conventional ways.
Beyond the text-video relationships, we further explore text-frame relationships with ambiguity-aware objectives since untrimmed videos often encompass multiple contexts.
Lastly, we employ cross-model ambiguity detection to mitigate the risk of error propagation in detecting ambiguous relationships that might occur when a model predicts and uses its prediction for training.

Our main contributions are:
(1) We propose ambiguity-restrained representation learning~(ARL), the first approach to address label ambiguity in PRVR. 
By modeling the relationships between instances, ARL mitigates the impact of learning uncertain relationships.
(2) We extend ARL to the text-frame level to account for multiple contexts in untrimmed videos, enhancing the effective utilization of all frames during learning.
(3) We introduce cross-model ambiguity detection to avoid learning erroneously detected ambiguity repeatedly.
(4) We achieves state-of-the-art performance on two datasets, \textit{i.e.}, TVR and ActivityNet.

\begin{figure*}[t]
    \centering
    \includegraphics[width=0.96\textwidth]{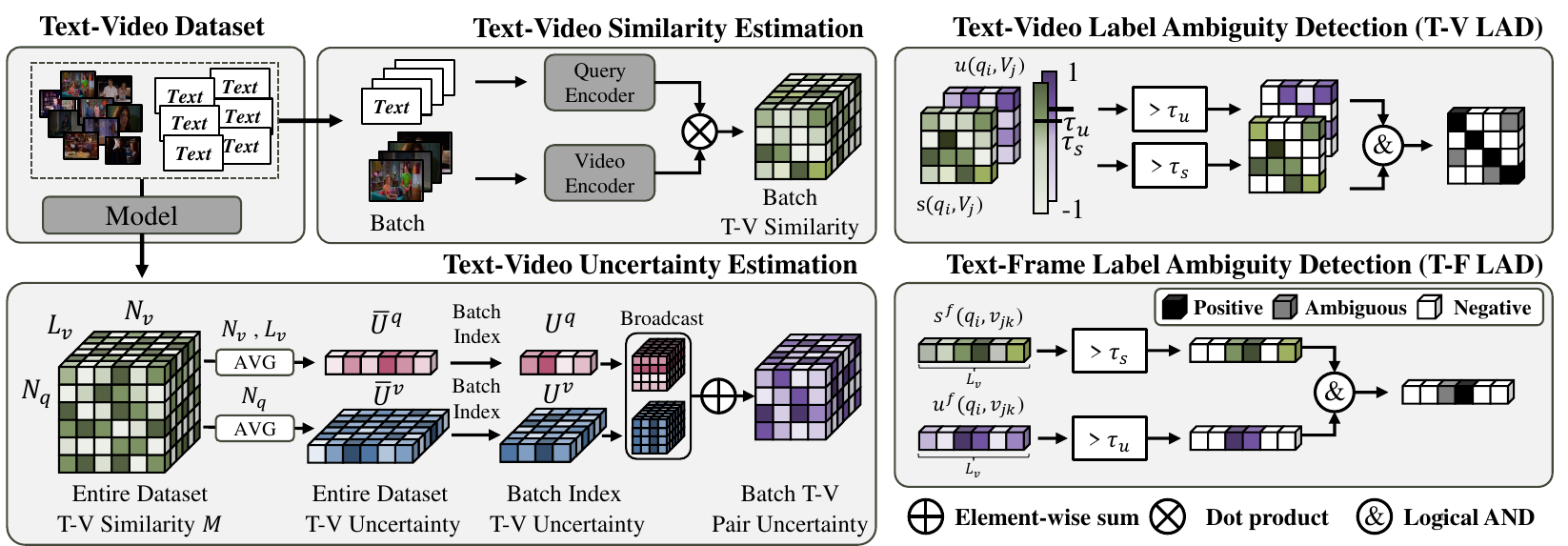}
    
    \caption{Overview of ARL.
     \textit{(Left)} Given the text-video train set, we initially calculate the text-video similarity to compute the uncertainty over the dataset level. 
     Batch-wisely indexed uncertainty is used along the batch-wise similarity between texts and videos to explore ambiguous text-video relationships in the mini-batch.  
    \textit{(Right)} Two levels of Label Ambiguity Detection~(LAD) modules detect ambiguous relationships. Text-Video LAD employs $s(q_i, V_j)$ and $u(q_i, V_j)$, the similarity and uncertainty maps between the text queries and videos in each mini-batch. 
    Text-Frame LAD utilizes $s^f(q_i, v_{jk})$ and $u^f(q_i, v_{jk})$, the similarity and uncertainty maps between each query and video frames for each text-video pair.}
    \label{fig:method_overview}
\end{figure*}

\section{Related Works}

\subsection{Text-to-Video Retrieval}
Text-to-video retrieval~(T2VR) aims to realize the metadata-free searching paradigm by aligning video contexts and text descriptions.
To benefit from the vision-language aligned model, the CLIP model pretrained on large-scale text-image pairs is popularly employed as the initial learning point.
Subsequently, tackling the mismatch between the amount of information in videos and texts, several works have focused on designing base units for feature matching~\cite{vrmis1, vrmis2}, e.g., frame-word~\cite{vrmis3} and frame-sentence~\cite{jin2023text, wu2023cap4video}.

Uncertainty in the scope of contexts has been discussed in the literature of T2VR.
Particularly, there have been approaches \cite{UATVR,PAU,song2019polysemous} to tackle the uncertainty.
PVSE~\cite{song2019polysemous} and UATVR~\cite{UATVR} extracted multi-faceted representations of text-video pairs.
PAU~\cite{PAU} addressed the aleatoric uncertainty inherent in text-video data by ensuring the consistency between different uncertainty measures. 
Our study shares a common high-level concept of tackling uncertainty in text-video data. 
Yet, our work differs in that the key focus is to explore ambiguous relationships among all text-video pairs.

\subsection{Partially Relevant Video Retrieval}

Beyond the scenario of T2VR, PRVR further targets the fine-grained capability of search engines to retrieve videos even when only the partial context corresponds to the given text queries.
To address PRVR, typical approaches are to clip the videos into multiple segments.
MS-SL~\cite{ms-sl} exhaustively constructed clips of varying lengths and performed similarity matching to the text query.
GMMFormer~\cite{wang2024gmmformer} implemented the feature of locality in forming clip representations by applying normally distributed weights in the attention layers.

On the other hand, our work has a key focus on detecting the ambiguous relationships between text-video instances caused by one-to-one labeling in video retrieval datasets.

\subsection{Noisy Label Learning}
Learning with noisy labels has been spotlighted due to its practical applications~\cite{Co-teaching,li2020dividemix,azadi2015auxiliary,wang2019symmetric}.
Uncertainty estimation and the co-training frameworks are the other popular streams.
Uncertainty is often utilized to detect noisy labels~\cite{neverova2019correlated,ju2022improving,northcutt2021confident,zheng2021rectifying} where co-training frameworks are shown to be effective in refining the noisy labels~\cite{Co-teaching,wei2020combating,tan2021co,li2020dividemix}.
This issue has also been addressed recently in the context of video-text learning~\cite{lin2024multi}.
To combat possible ambiguity in pair-wisely labeled relationships, our study employs the concept of uncertainty and co-training.

\section{Method}

\subsection{Overview}
\label{sec:overview}

The overview of Ambiguity-Restrained representation Learning~(ARL) is shown in Fig.~\ref{fig:method_overview}. 
To illustrate, we compute the similarity between all texts and videos in the training set to define text-video uncertainty in each epoch.
Subsequently, the batch-wisely indexed uncertainty is processed to the Label Ambiguity Detection~(LAD) module along with the similarity between the texts and videos in the mini-batch.
LAD identifies ambiguous relationships between text and video modalities at two levels: the text-video and text-frame. 
Lastly, we perform cross-model ambiguity detection with the dual branch structure which is commonly employed in the literature of PRVR.
Employing the same structure and inputs for each branch for cross-model ambiguity detection, cross-model ambiguity detection allows each model to learn with the detected ambiguous sets from another model.
Note that the cross-model ambiguity detection is not depicted in Fig.\ref{fig:method_overview}.
For the rest of the paper, we explain the training scenario with an assumption that all text queries and video instances are composed of the same number of elements, i.e., $L_q$ words and $L_v$ frames, for better clarity.

\paragraph{Text Query Representation.} 
Given $i$-th text query out of $N_q$ queries in the training dataset, we use the pretrained text encoder to extract features for each word.  
Subsequently, we embed the word features into a low-dimensional space using a fully connected (FC) layer followed by ReLU activation. 
After that, we incorporate positional encoding into these features and employ a transformer layer to obtain $d$-dimensional word feature vectors~$Q_i \in \mathbb{R}^{L_{q} \times d}$. Finally, we apply an attention pooling module to the word feature vectors to obtain the query text embeddings $q_i \in \mathbb{R}^{d}$.

\paragraph{Video Representation.}
Given $j$-th untrimmed video out of $N_v$ train set videos, we extract frame features $V_j' \in \mathbb{R}^{L_v \times d_v}$ using a pre-trained 2D or 3D CNN. 
Symmetric to the text branch, video features are also passed through a fully connected layer with ReLU activation to reduce the dimensions. 
After that, we incorporate positional encoding $P$ into the extracted features, before forwarding the features through a transformer layer to obtain $V_j \in \mathbb{R}^{L_{v} \times d}$:
\begin{align}
    \!\!V_j\!=\! [ v_{j1},v_{j2}, ...,v_{jL_{v}} ] \!=\! \text{Transformer}(\text{FC}(V_j')+P),
\end{align}
where $v_{jk}$ refers to the $k$-th frame features in the $j$-th video.  

\paragraph{Similarity Measure.}
Given text and video representations, the frame-wise similarity score~$s^f$ between a text query features~$q_i$ and video frame features~$v_{jk}$ is derived as:
\begin{align}
\label{Eq.intra_sim}
    s^f(q_i, v_{jk}) = \text{cos}(q_i,v_{jk}),
\end{align}
where cos$(\cdot, \cdot)$ is the cosine similarity between text query and video frames. 
Then, the maximum similarity value between a text query and video frames is used for the retrieval score since only partial video frames are relevant to the text.
The retrieval (similarity) score is obtained as follows:
\begin{align}
\label{Eq.similaritymeasure}
    s(q_i, V_j) = s^f(q_i, v_{j\hat{k}}),
\end{align}
where $v_{ j \hat{k} } = \text{argmax}_{ v_{jk} } \text{cos}( q_i, v_{jk} )$ and $\hat{k}$ denote the video frame 
 and its index with the maximum similarity, repectively.

\subsection{Ambiguity-Restrained Representation Learning}
\label{sec:ambiguity}
\textbf{Definition of Ambiguity.} Typical datasets for Partially Relevant Video Retrieval (PRVR) consist of thousands of matched text-video pairs. 
It is a popular practice to treat paired text-video data as positive pairs, while all unpaired data are regarded as negative pairs. 
However, we question whether it is correct to treat all unpaired data as negative.

We address this issue by defining an ambiguous relationship for unpaired text-video pairs that are difficult to classify as simply negative. 
Specifically, we use uncertainty and similarity measures to identify these ambiguous relationships and grant the model flexibility in treating these instances.
First, the uncertainty indicates whether each instance contains commonly shared semantics.
It is calculated as the average similarity across the dataset, and thus, the uncertainty is measured high if numerous instances possess high similarity to a specific instance.
Second, similarity refers to the degree of pair-wise text-video resemblance, defined as the maximum similarity between a single text query and video frame representations.
Thus, the pairs with high similarity indicate that a video instance includes a frame that shares similar attributes with the given query.
Although both metrics originate from the same similar operation, we note that these two metrics serve as different criteria with different objectives.
We also note that our study in Fig.~\ref{fig:pos_neg_s_u} further validates the claim that these two metrics exhibit different distributions.

\paragraph{Warmup Phase.}
Our method utilizes online knowledge~(per epoch) of the model to identify the ambiguous relation between text and video.
Therefore, we need to warm up the model for a few epochs to initially train the model to learn general text-video relationships.
For the warmup phase, we adopt the commonly used triplet ranking loss~\cite{dldkd,vse++} and infoNCE loss~\cite{X-clip_VR,reloclnet} in retrieval tasks.

\paragraph{Uncertainty Estimation.}

Uncertainty for each text/video instance is measured by considering similarities across the entire dataset before training progress.
Specifically, we calculate the feature similarity map~$M \in \mathbb{R}^{N_{q}\times N_{v} \times L_{v}}$ between all text queries and video frames in the training dataset using the online model.
Here, \( M_{xyz} \) denotes the similarity between the \( x \)-th text query and the \( z \)-th frame of the \( y \)-th video in the entire dataset.
With the similarity map $M$, we define the dataset-wise uncertainty of each text query~$\bar{U}^{q} \in \mathbb{R}^{N_q}$ as the average similarity between the text and all video frames in the dataset and the uncertainty of video frame~$\bar{U}^{v} \in \mathbb{R}^{N_v \times L_v}$ as the average similarity to all text queries:
\begin{align}
\label{Eq.uncertainty}
    \bar{U}_{x}^{q} = \frac{1}{N_{v}L_{v}} \sum_{y=1}^{N_{v}} \sum_{z=1}^{L_{v}} M_{xyz},  \quad  \bar{U}_{yz}^{v} = \frac{1}{N_{q}} \sum_{x=1}^{N_{q}}M_{xyz},
\end{align}
where $\bar{U}_{x}^{q}$ is uncertainty value of $x$-th text query and $\bar{U}_{yz}^{v}$ is uncertainty value of $z$-th frame of $y$-th video instance.
Texts with higher average similarity to video frames and frames with higher average similarity to queries both exhibit greater uncertainty.

Note that an instance with higher uncertainty implies that its context is likely to be commonly shared with other instances.
These similarity maps and uncertainties are updated with Eq.~\ref{Eq.uncertainty} every epoch.

For better clarity for indexing uncertainties $U^q$ and $U^v$ within each mini-batch, we define the batch-level subsets of uncertainties as $U^q$ and $U^v$.
Note that $U^q$ and $U^v$ are subsets of $\bar{U}^q$ and $\bar{U}^v$, which include the uncertainties of all queries and videos in the mini-batch.

Below, we define the uncertainty value between the text query and each video frame as \(u^f\) and the uncertainty between the query and the whole video as \(u\):

\begin{align}
\label{Eq.uncertainty2}
\!\!\!u(q_i,V_j)\!=\!\frac{1}{2}({U_{i}^{q}}+{U_{j\hat{k}}^{v})};
    u^f(q_{i},v_{jk})\!=\!\frac{1}{2}({U_{i}^{q}}+{U_{jk}^{v})} 
    ,
\end{align}
where $\hat{k}$ indicates the frame index with the highest similarity to $i$-th text query, as mentioned in Eq.~\ref{Eq.similaritymeasure}.

\paragraph{Text-Video Label Ambiguity Detection.}

To discover pairs in ambiguous relation per text~/~video instance, we utilize computed uncertainty~$u$ and similarity score~$s$.  For $i$-th text query $q_i$, a set of ambiguous video pairs are gathered as:
\begin{align}
\label{Eq.detect_A_q}
    \mathcal{A}_i^q = \left\{ V_a \mid s(q_i, V_a) > \tau_s \ \text{and} \ u(q_i, V_a) > \tau_u \right\},
\end{align}
where $\tau_s$ and $\tau_u$ are thresholding hyperparameters.

On the other hand, an ambiguous query set for a video~$V_j$ is defined as follows:
\begin{align}
\label{Eq.detect_A_v}
    \mathcal{A}_j^v = \left\{ q_a \mid s(q_a, V_j) > \tau_s \ \text{and} \ u(q_a, V_j) > \tau_u \right\}.
\end{align}
\paragraph{Ambiguous Aware Representation Learning.}

For training, we utilize margin triplet ranking loss and contrastive learning~\cite{simclr}, following previous works~\cite{ms-sl, wang2024gmmformer}.
Below, we enumerate the modifications in objectives to implement with ambiguous relationships.
In the case of contrastive learning, we add modifications to supervised contrastive learning~\cite{supcon} to equip with the multi-positive contrastive objective:
\begin{align}
\label{Eq.contrastive_t2v}
\mathcal{L}^\text{{t2v}}_{ij} =
     -\log\left( \frac{e^{s(q_i,V_j)}+\sum_{V_{a}\in \mathcal{A}_{i}^{q}} e^{s(q_i,V_a)}}
    {e^{s(q_i,V_j)}+\sum_{V \in 
 \mathcal{A}_{i}^{q} \lor\mathcal{N}^q_{i}}e^{s(q_{i},V)}} 
 \right)
 \end{align}
 \begin{align}
 \label{Eq.contrastive_v2t}
\mathcal{L}^\text{{v2t}}_{ij} = -\log\left( \frac{e^{s(q_i,V_j)}+\sum_{q_{a}\in \mathcal{A}_{j}^{v}} e^{s(q_a,V_j)}}
    {e^{s(q_i,V_j)}+\sum_{q\in \mathcal{A}_{j}^{v} \lor \mathcal{N}^v_{j}}e^{s(q,V_j)}}  \right)
\end{align}
 \begin{align}
\mathcal{L}^\text{{nce}} = \frac{1}{n} \sum_{(q_i,V_j)\in\mathcal{B}} 
\mathcal{L}^\text{{t2v}}_{ij} +
\mathcal{L}^\text{{v2t}}_{ij},   
 \end{align}
where \(\mathcal{B}\) denotes a mini-batch, and (\(q_i\), \(V_j\)) denotes a positive pairs within this batch. 
$\mathcal{N}^v_{j}$ and $\mathcal{N}^q_{i}$ are sets of negative samples for each video and query that $\mathcal{N}^v_{j}$ contains samples that are neither in positive nor ambiguous relations to $j$-th video.
In short, our multi-positive contrastive objective enables more flexible model learning by accommodating ambiguous relationships. 
While instances within ambiguous sets in the numerator are not trained as negatives, not all instances are necessarily trained to have positive relations with the anchor. 
This is because of the possibility that maximizing a single similarity value can still facilitate loss convergence.

On the other hand, we organize dual triplets for margin triplet ranking loss; one with ambiguous sets for $\mathcal{L}_{a}^\text{trip}$ and the other with negative pairs for $\mathcal{L}_{n}^\text{trip}$ as follows:
\begingroup
\footnotesize
\begin{multline}
    \mathcal{L}_{a}^\text{trip} = \frac{1}{n} \sum_{(q_i,V_j)\in\mathcal{B}}\{ \max(0,m_a + s(q_a,V_j) - s(q_i,V_j)) +\\ \max(0,m_a + s(q_i,V_a) - s(q_i,V_j)) \}
    \label{Eq.triplet}    
\end{multline}
\begin{multline}
    \mathcal{L}_{n}^\text{trip} = \frac{1}{n} \sum_{(q_i,V_j)\in\mathcal{B}}\{ \max(0,m + s(q_n,V_j) - s(q_i,V_j)) + \\ \max(0,m + s(q_i,V_n) - s(q_i,V_j)) \},
    \label{Eq.triplet_2}    
\end{multline}
\endgroup
where \(m_a\) and \(m\) denote the respective margins, and \(q_a \in \mathcal{A}_j^v\) and \(V_a \in \mathcal{A}_i^q\) represent the ambiguous samples for each video and query, respectively.
$q_n \text{ and } V_n$ are the negative samples for each query and video.
We set the margin for \(\mathcal{L}_{a}^\text{trip}\) smaller, ensuring that ambiguous instances remain more similar to the anchor than contextually irrelevant instances (\(m_a < m\)).
By alleviating distance constraints on ambiguous instances, we allow ambiguous sets with potential positive relationships to the anchor to be excluded from negative training. The margin $m_a$  is used to maintain the hierarchy, ensuring that paired instances are trained as positives in Eq.~\ref{Eq.contrastive_t2v}-~\ref{Eq.contrastive_v2t}. 
In conclusion, the ambiguity-restrained objective for text-video pairs is formulated as follows: $\mathcal{L}^{\text{video}}=\lambda_{\text{nce}}\mathcal{L}^{\text{nce}}+\mathcal{L}_{a}^{\text{trip}}+\mathcal{L}_{n}^{\text{trip}}$, where $\lambda_{\text{nce}}$ is a hyperparameter to balance the losses.

\paragraph{Text-Frame Label Ambiguity Detection Within Untrimmed Video.}

 \label{tf_lad}

For untrimmed videos, diverse contexts may be present within a single instance.
Yet, exploring the relationships with the same video is a yet-to-be-explored issue.
Therefore, we delve deep into the relation between the text query and frame-wise representations.
Symmetrical to the process of discovering the ambiguity relations between text and video instances in the mini-batch, we apply Eq.~\ref{Eq.detect_A_q} and Eq.~\ref{Eq.detect_A_v} between the query feature $q$ and video frame feature $v$.
Note that the similarity and uncertainty measures, \textit{i.e.,} $s$ and $u$, are also substituted with frame-wise similarity $s^f$~(Eq.~\ref{Eq.intra_sim}) and uncertainty $u^f$~(Eq.~\ref{Eq.uncertainty2}), respectively.
Consequently, the same objectives for learning text-video relationships in Eq.~\ref{Eq.contrastive_t2v}-~\ref{Eq.triplet_2} are applied to learn text-frame relationships for each video. 
The objective for text-frame pairs $\mathcal{L}^{\text{frame}}$ is also the same with $\mathcal{L}^{\text{video}}$.
We note that $\lambda_{\text{nce}}$ is shared.

\subsection{Cross-Model Ambiguity Detection}
\label{Sec.coteaching}

{Ambiguous pairs can be detected using the model's own predictions, similar to self-training~\cite{self_training_1,self_training_2}.
However, we point out the vulnerability of error propagation of ambiguity detection when the model relies on its knowledge, progressively reinforcing its initial imperfect predictions.

To address this challenge, we utilize two identical encoders that reciprocally transfer one's detected ambiguous sets to the other to mitigate the impact of noisy labels~\cite{Co-teaching}.
Given two models denoted as $\theta$ and $\Phi$, each model computes ambiguous text-video pairs following Eq.~\ref{Eq.uncertainty}-Eq.~\ref{Eq.detect_A_v} with its online knowledge and provides it as the training guidance to the other model. 

Finally, the retrieval score for the prediction is yielded as the average of Eq.~\ref{Eq.similaritymeasure} for each model:
\begin{align}
    s(q_i,V_j) = \frac{1}{2} \big( s_{\theta}(q_i,V_j) + s_{\Phi}(q_i,V_j) \big),
\end{align}
where $s_{\theta}$ and $s_{\Phi}$ denote the retrieval scores from model $\theta$ and $\Phi$, respectively.

\section{Experiments}
\begingroup
\setlength{\tabcolsep}{5pt} 
\renewcommand{\arraystretch}{0.96} 
\begin{table*}[t]
\centering
\small
{

 \begin{tabular}{l|ccccc|ccccc}
 \hlineB{2.5}
 \multirow{2}{*}{Model} & \multicolumn{5}{c|}{TVR} & \multicolumn{5}{|c}{ActivityNet Captions} \\
 \cline{2-11}
& R@1 & R@5 & R@10 & R@100 & SumR & R@1 & R@5 & R@10 & R@100 & SumR  \\ \hlineB{2.5}
\rowcolor{gray!20}\multicolumn{11}{l}{\textit{VCMR methods without moment localization}} \\ 
\hline
XML~\cite{tvr}  & 10.0& 26.5 & 37.3 & 81.3 & 155.1 & 5.3 & 19.4 & 30.6 & 73.1  & 128.4\\
ReLoCLNet~\cite{reloclnet} & 10.7 & 28.1 & 38.1 & 80.3 & 157.1 & 5.7 & 18.9 & 30.0 & 72.0 & 126.6\\
CONQUER~\cite{conquer}& 11.0 & 28.9 & 39.6 & 81.3 & 160.8 & 6.5 & 20.4 & 31.8 & 74.3  & 133.1\\
\hline
\rowcolor{gray!20}\multicolumn{11}{l}{\textit{PRVR models}} \\ 
\hline
MS-SL~\cite{ms-sl} & 13.5 & 32.1 & 43.4 & 83.4 & 172.4  & 7.1 & 22.5 & 34.7 & 75.8  & 140.1 \\
\color{gray} \color{gray}DL-DKD~\cite{dldkd}$\dagger$ & \color{gray}14.4 & \color{gray}34.9 & \color{gray}45.8 & \color{gray}84.9 & \color{gray}179.9 & \color{gray}8.0 & \color{gray}\textbf{25.0} & \color{gray}\textbf{37.4} & \color{gray}77.1  & \color{gray}147.6 \\ 
GMMFormer~\cite{wang2024gmmformer} & 13.9 & 33.3 & 44.5 & 84.9 & 176.6 & \textbf{8.3} & 24.9 & 36.7 & 76.1  & 146.0\\ 
\hline
 \textbf{Ours} & \textbf{15.6} & \textbf{36.3} & \textbf{47.7} & \textbf{86.3} & \textbf{185.9} & \textbf{8.3} & 24.6 & \textbf{37.4} & \textbf{78.0} & \textbf{148.3} \\ \hlineB{2.5}

 \end{tabular}
 \caption{Performance comparison for Resnet, I3D and Roberta features. $\dagger$ indicates the usage of additional CLIP-B/32 model.}
 \label{table_i3d}
}
\end{table*}
\endgroup

\begingroup
\setlength{\tabcolsep}{6pt} 
\renewcommand{\arraystretch}{0.96} 
\begin{table*}[t]
\centering
\small
{

 \begin{tabular}{l|ccccc|ccccc}
 \hlineB{2.5}
 \multirow{2}{*}{Model}&\multicolumn{5}{c|}{TVR}&\multicolumn{5}{|c}{ActivityNet Captions}\\
 \cline{2-11}
& R@1 & R@5 & R@10 & R@100 & SumR & R@1 & R@5 & R@10 & R@100 & SumR  \\ \hlineB{2.5}

MS-SL~\cite{ms-sl}& 31.9 & 57.6 & 67.7 & 93.8 & 251.0& 14.7 & 37.1 & 50.4 & 84.6 & 186.7 \\
GMMFormer~\cite{wang2024gmmformer} & 29.8 & 54.2 & 64.6 & 92.5 & 241.1 & 15.2 & 37.7& 50.5 & 83.7  & 187.1\\ 
\hline
\textbf{Ours} & \textbf{34.6} & \textbf{60.4} & \textbf{70.7} & \textbf{94.4} & \textbf{260.1} & \textbf{15.3} & \textbf{38.4} & \textbf{51.5} & \textbf{85.2} & \textbf{190.4} \\ \hlineB{2.5}

 \end{tabular}
 \caption{Performance comparison on TVR and ActivityNet Captions for CLIP-L/14 features.}
 \label{table_clip}
}

\end{table*}
\endgroup

\subsection{Datasets and Metrics}

We evaluate our method on two large-scale video datasets,  \textit{i.e.} TVR~\cite{tvr} and ActivityNet Captions~\cite{anetcaptions}. 
We adopt the data split provided by previous works~\cite{zhang2020hierarchical,reloclnet}. 
Following~\cite{ms-sl}, we use rank-based recall as our evaluation metric, \textit{i.e.,} R@K (K=1, 5, 10, 100), where R@K denotes the fraction of queries that successfully retrieve the desired items within the top K of the ranking list. 
Moreover, we report the Sum of all Recalls (SumR) for comprehensive comparisons.

\subsection{Implementation Details}
\label{implementation}

We use ResNet~\cite{resnet} and I3D~\cite{i3d} for TVR and only the I3D  for the ActivityNet-Captions to extract visual features. 
For text query representation, we use the RoBERTa~\cite{liu2019roberta} feature for both datasets. 
Also, to demonstrate the effectiveness of our method with the large model, we conducted experiments using CLIP-L/14~\cite{clip}. 
While the typical parameters are set the same as in \cite{wang2024gmmformer}, the thresholds $\tau_{s}$ and $\tau_{u}$ were defined at each epoch using the distribution values of similarity and uncertainty from the training dataset. 
Particularly, $\tau_{s}$ is set to the mean value of the similarity distribution of the positive pairs, and $\tau_{u}$ is set to the value corresponding to the mean of uncertainty distribution of train dataset. 
More details are provided in the appendix.

\subsection{Comparison With the State-of-the-Arts}

\noindent\textbf{Retrieval Performance.}
We report the results comparing our approach with state-of-the-art methods for video corpus moment retrieval, and PRVR.
Note that all performances are yielded without using the moment supervision.
In Tab.~\ref{table_i3d} and Tab.~\ref{table_clip}, we show the results on the TVR and ActivityNet Captions datasets.
As observed, our proposed method outperforms previous works in all recall metrics, achieving a margin of 9.3\%, 2.3\% in SumR compared to GMMFormer when using ResNet, I3D, and Roberta. 
Moreover, we highlight the consistent gaps between ours and DL-DKD~\cite{dldkd} while DL-DKD employed an additional CLIP-B/32 model for knowledge distillation~(marked with $\dagger$). 
We owe this superior performance to the study of ambiguity in one-to-one relationship learning with untrimmed videos for PRVR.
This trend was similarly observed with CLIP-L/14, showing improvements of 9.1\% and 3.3\% compared to MS-SL and GMMFormer respectively.

\noindent\textbf{Complexity Analysis.}
In this section, we present the model complexity analysis, as shown in Tab \ref{table_flops_params} and \ref{table_complexity}. 
We analyzed three aspects: FLOPs, the number of parameters, and runtime required to process a single text query on Nvidia RTX 3090 GPU. 
While our method demonstrates high efficiency in terms of FLOPs and parameters among PRVR methods, its runtime is relatively slower than that of GMMFormer.
The efficiency in FLOPs and the number of parameters is attributed to our streamlined transformer architecture, whereas GMMFormer employs multiple attention blocks in parallel with different Gaussian kernels. 
The relatively slower runtime of our method is due to the absence of aggregated frame-level feature used in GMMFormer. 
However, our method still achieves a real-time runtime of under 1 ms, which represents a well-balanced tradeoff between performance and runtime speed.

\begin{figure*}[t]
    \begin{center}
        \noindent\includegraphics[width=0.9\textwidth]{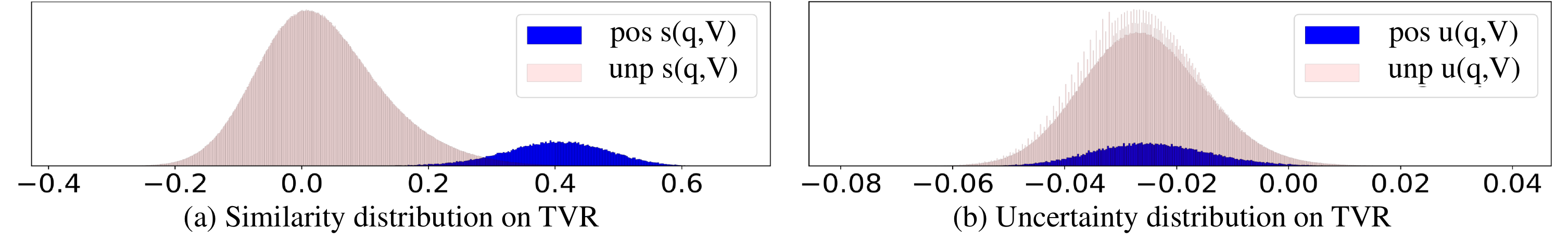}
    \end{center}    
    \caption{
        Similarity and uncertainty distributions for positively-paired and unpaired text-video pairs on TVR. 
        Similarity distributions are illustrated in (a).
        Distributions of positive sets are generally higher than that of negative sets.
        In (b), uncertainty distributions are shown.
        As uncertainty value is not much affected by the similarity of single pair~(for positive pairs), it is shown that distributions of positive and unpaired are formed similarly.
        }
        \label{fig:pos_neg_s_u}
\end{figure*}

\begin{figure*}[t]
    \begin{center}
        \noindent\includegraphics[width=0.95\textwidth]{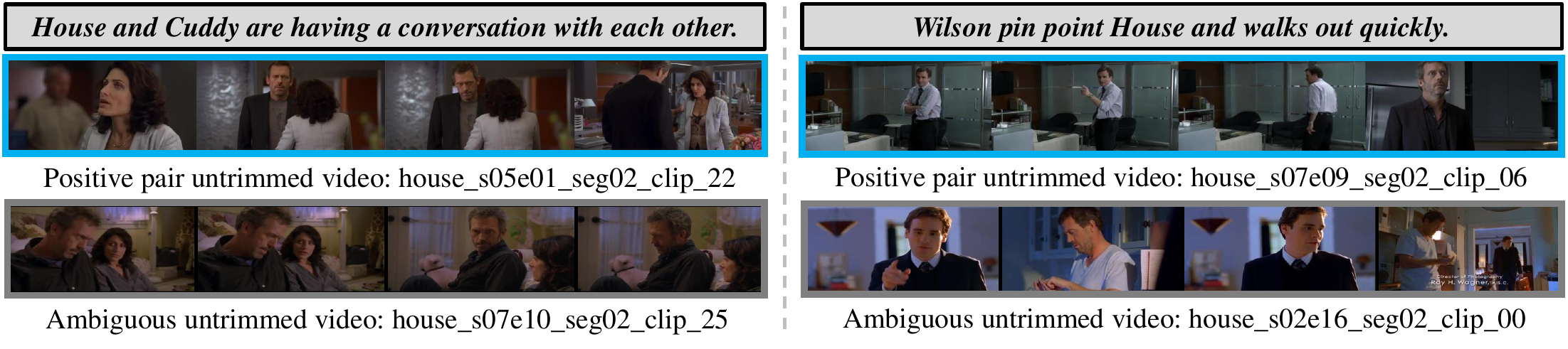}
    \end{center}    
    \caption{ Ambiguity detection results on TVR dataset.
For given queries, we visualize untrimmed videos that are detected to be in ambiguous relations after the training process (gray boxes).
The videos are shown to be highly relevant to the query although they are not paired as positive instances of the query.}
    \label{fig:ambiguous_sample}
\end{figure*}

\begin{table}

\centering
\footnotesize
{
\begin{tabular}{l|c|c|c}
\hlineB{2.5}
 & MS-SL & GMMFormer & Ours    \\ \hline

FLOPs (G) &  1.29 &  1.95&  1.23   \\
Params (M) &  4.85 &  12.85 & 5.34     \\ 

\hlineB{2.5}
\end{tabular}
\caption{Model complexity comparison.}
\label{table_flops_params}
}

\end{table}

\begin{table}

\centering
\footnotesize
{
\begin{tabular}{l|cccc}
\hlineB{2.5}
\rowcolor{gray!20}\multicolumn{5}{l}{\textit{inference runtime (ms)}} \\ 
\hline
Video size & 1000 & 1500 & 2000 & 2500    \\ \hline

MS-SL & 0.366 & 0.606 & 0.759 & 0.893    \\
GMMFormer & 0.264 & 0.267 & 0.270 & 0.293      \\ 

Ours & 0.294 & 0.391 & 0.427 & 0.612      \\ \hlineB{2.5}
\end{tabular}
\caption{Inference time comparison.}
\label{table_complexity}
}

\end{table}

\subsection{Ablation and Further Studies}
\noindent\textbf{Component Analysis.}
To understand the effectiveness of each component, we conduct component ablation in Tab.~\ref{table_ablation}.
Comparing rows (a) and (b), our base model and the model trained with text-video ambiguity in representation learning, we observe that it is effective in that it results in a 3.6-point increase in SumR.
It demonstrates that the flexibility in learning text-video relationships is important for PRVR because text queries and videos often exhibit ambiguity in contextual similarity.
Moreover, the increase in rows (c) and (e) spotlight the potential of exploring the text-frame relationships within each video.
Subsequently, cross-model learning is verified to be beneficial for PRVR as shown in rows (d) and (e). To conclude, the benefits of our components imply that considering only the paired text-video instances as contextually similar to each other in pair-wisely labeled datasets is vulnerable to forming ambiguous relationships between text-video instances~(especially for PRVR).

\begingroup
\setlength{\tabcolsep}{4pt} 
\renewcommand{\arraystretch}{0.96} 
\begin{table}[t]
\centering
{
 \small

 \begin{tabular}{l|ccc|ccccc}
 \hlineB{2.5}
  & T-V & T-F & C.L &  R@1 & R@5 & R@10 & R@100 & SumR \\ \hline
 (a) & - & - & - & 32.8 & 58.1 & 68.2 & 93.7 & 252.8  \\ 
 (b) &\checkmark & - & - & 33.6 & 58.9 & 69.4 & 94.5 & 256.4  \\ 
 (c) &\checkmark & \checkmark & - & 34.3 & 59.9 & 70.1 & 94.4 & 258.7 \\
 (d) &\checkmark & - & \checkmark & 34.3 & 59.9 & 69.9 & 94.3 & 258.4\\ 
 (e) &\checkmark & \checkmark & \checkmark  & 34.6 & 60.4 & 70.7 & 94.4 & 260.1 \\ 
 \hlineB{2.5}
 \end{tabular}
 \caption{Ablation study to investigate the effectiveness of different components on TVR. T-V Ambiguity and T-F Ambiguity indicate the use of ambiguity-aware representation learning within text-video representation learning and the use of text-frame representation learning.}
 \label{table_ablation}
}

\end{table}
\endgroup

\begin{table}[t]

\centering
\footnotesize
{
\begin{tabular}{l|ccccc}
\hlineB{2.5}
$\mathcal{A}$ & R@1 & R@5 & R@10 & R@100 & SumR \\ \hline
Positive & 34.0 & 59.7 & 70.1 & 94.5 & 258.3\\
Ignore & 34.5 & 60.1 & 70.1 & 94.5 & 259.2\\
Ours & 34.6 & 60.4 & 70.7 & 94.4 & 260.1  \\ 
\hlineB{2.5}
\end{tabular}
\caption{Performances comparison between different usages of ambiguous sets.
\label{table_limitation}
}

}

\end{table}

\noindent\textbf{Ambiguous Set Learning Strategy.}
In our work, we granted flexibility to the model in learning ambiguous relationships.
Yet, other options exist to treat all text-video instances in ambiguous sets as positives for a given anchor or to exclude them from training.
In Tab.~\ref{table_limitation}, we report the performances of other options.
Specifically, we employed a supervised contrastive objective~\cite{supcon} to maximize the similarity between every ambiguous pair for positive or utilized masking operation to ignore.
As the performances decrease in both scenarios, we believe that ambiguous sets $\mathcal{A}$ are the mixture of instances in positive and negative relations with an anchor. 
Consequently, these results indicate that simply defining an ambiguous set as positive or negative could lead to suboptimal results.

\noindent\textbf{Uncertainty \& Similarity Distributions.}
 we argued that the two metrics used to detect ambiguity serve different roles and exhibit distinct distributions. 
In Fig.~\ref{fig:pos_neg_s_u}, we present the distributions of each metric on the TVR. (ActivityNet Captions is similar to the TVR.)
As illustrated, the distributions of similarity~(left) and uncertainty~(right) display different shapes. 
Specifically, the positive and negative distributions are distinguishable in terms of similarity, indicating that similarity between positively paired pairs is generally higher than that of unpaired text-video pairs. 
Conversely, we observe that uncertainty values for positive and unpaired sets are similarly distributed since the degree of semantic overlap throughout the dataset does not depend on a single pair-wise similarity.
Note that the text-video averaged uncertainty values are used for plots as pairwise uncertainty is utilized to detect ambiguous pairs~(Eq.~\ref{Eq.uncertainty2}).
These results indicate that similarity and uncertainty hold independent meanings that both are preferred to be considered simultaneously.

\subsection{Qualitative Results}
We analyzed to verify whether detected text-video pairs actually include ambiguous relationships to an anchor. 
In Fig.~\ref{fig:ambiguous_sample}, we plot two text queries with the paired videos and the videos in ambiguous relationships~(with gray boxes).
To illustrate, we observe that the characters in the scenes are identical while the overall context is also very similar.
Especially, the video on the left example includes the moment that can be expressed with the given query.
This confirms that our ambiguity detection effectively captures ambiguous relationships and reduces the impact of erroneous supervision that occurs when treating all unpaired text-video pairs as negative sets.

\section{Conclusion}
\label{conclusion}

\paragraph{Conclusion.}
In this paper, we addressed the issue of ambiguous relationships between the text-video pairs in pairwise labeled text-video data. 
To tackle this challenge, we proposed Ambiguity-Restrained representation Learning (ARL), designed to mitigate the impact of learning from ambiguous relationships between text and video. 
ARL utilizes uncertainty to first assess whether each text or video likely includes common contexts within the dataset. 
Subsequently, the similarity within each mini-batch is computed to identify ambiguous relationships. 
These relationships are then incorporated into the Ambiguity-Aware Representation Learning framework, allowing the model flexibility in learning these relationships.
Our results demonstrate that one-to-one relationship learning is vulnerable to ambiguous relationships between texts and videos.

\section{Acknowledgments}
This work was supported in part by MSIT/IITP (No. 2022-0-00680, 2020-0-01821, 2019-0-00421, RS-2024-00459618, RS-2024-00360227, RS-2024-00437102, RS-2024-00437633), and MSIT/NRF (No. RS-2024-00357729).

\bibliography{aaai25}

\end{document}